\documentclass[a4paper]{article}

\usepackage{INTERSPEECH2021}
\usepackage{caption}
\usepackage{subcaption}
\usepackage{float}
\usepackage{amssymb}
\usepackage[table,xcdraw]{xcolor}
\usepackage{svg}

\newcommand{\etal}{\textit{et al}. }

\title{Collaborative
Training of Acoustic Encoders for Speech Recognition}
\name{\centering Varun Nagaraja\textsuperscript{*}\thanks{* Equal Contribution}, Yangyang Shi\textsuperscript{*}, Ganesh Venkatesh, Ozlem Kalinli, Michael L. Seltzer, \newline Vikas Chandra}
\address{Facebook, USA}
\email{\{vnagaraja, yyshi\}@fb.com}

\begin{document}

\maketitle
\begin{abstract}
On-device speech recognition requires training models of different sizes for deploying on devices with various computational budgets.
When building such different models, we can benefit from training them jointly to take advantage of the knowledge shared between them.
Joint training is also efficient since it reduces the redundancy in the training procedure's data handling operations.
We propose a method for collaboratively training acoustic encoders of different sizes for speech recognition.
We use a sequence transducer setup where different acoustic encoders share a common predictor and joiner modules.
The acoustic encoders are also trained using co-distillation through an auxiliary task for frame level chenone prediction, along with the transducer loss. 
We perform experiments using the LibriSpeech corpus and demonstrate that the collaboratively trained acoustic encoders can provide up to a 11\% relative improvement in the word error rate on both the test partitions.

\end{abstract}

\section{Introduction}
Speech interfaces are accessible today on many edge devices (e.g., mobile phone and smart speaker), which possess very different computational capabilities. Hence, it has become necessary to build speech recognition models of different sizes to accommodate the wide variety of computational resources. Many automatic speech recognition models \cite{Wu2020,shi2020emformer,gulati2020conformer} usually are proposed in a few different size configurations, which fall into the relative categories of small, medium, and large. Typically, the models of different sizes are trained independently of each other. This leads to inefficiency in the training steps due to repeated data loading and manipulation. The separately trained models might also be sub-optimal in performance since there is no knowledge shared between them.

One of the popular approaches for sharing knowledge between the models is Knowledge Distillation \cite{Hinton_KD} (KD). The traditional KD technique is a two-step procedure that trains a teacher model in the first step and uses the teacher's supervision to train a student model in the second step. While this procedure can help in improving the model's performance, it increases the total time for building a model due to the serialized training procedure. 
The serialization overhead can be avoided using a single-step version of knowledge distillation referred to as co-distillation \cite{zhang2018deep,anil18codistillation}. Co-distillation trains models jointly and has improved the performance and training efficiency of image classifiers \cite{lan2018knowledge}. We propose a method that applies co-distillation to train a group of speech recognition models of different sizes. 

A widely used on-device automatic speech recognition (ASR) model is the sequence transducer network \cite{graves2012sequence}. A sequence transducer model consists of an acoustic encoder, a predictor, and a joiner network. The commonly used acoustic encoders are multi-layered RNN/LSTM \cite{rao2017asru,he2018streaming} and Transformers \cite{Wu2020,shi2020emformer,gulati2020conformer,zhang2020transformer}. In this work, we use the low latency streaming Transformer \cite{vaswani_attention} based encoder proposed in the Emformer \cite{shi2020emformer} architecture. 

In the sequence transducer network, the acoustic encoder is the major contributor to the model's size and the computation cost. The acoustic encoder has a deeper structure than the predictor and the joiner. Hence, we design an on-device ASR model to have a configurable acoustic encoder with a shared predictor and a joiner network. A group of acoustic encoders are trained jointly in a single sequence transducer model. During deployment, an appropriate size encoder can be selected depending on the computational constraints. 

We construct a group of acoustic encoders of different sizes by using different number of layers. The encoders share a set of low-level layers and branch off into additional layers according to the size requirements. 
An auxiliary task projects the last layer outputs from each branch to logits used for frame-level chenone \cite{le2019senones} prediction. Such an auxiliary task was shown to be helpful by Liu \etal \cite{liu2020auxtasks} for improving acoustic encoders in sequence transducer model. Along with the benefit of training for the auxiliary task in each branch, we apply a KL divergence loss on the output probabilities, which helps in knowledge distillation between the encoders. 

Our proposed collaborative training method is a novel framework for single-step training of acoustic encoders of different sizes in a speech recognition model.
We demonstrate on the LibriSpeech dataset \cite{librispeech} that the collaboratively trained models can provide a 3-11\% relative improvement in the word error rate when compared to the models trained separately.
The joint training also helps in sharing the computational resources since we can perform operations, e.g., data loading and manipulation, just once for all the models. 

\section{Related work}
Knowledge distillation (KD) techniques have been used in the context of speech recognition for model compression \cite{Chebotar16,Pang18,panchapagesan2020efficient}, domain adaptation \cite{li2017KD,Asami17,Manohar18,Meng19,mosner2019KD} and transferring knowledge from full-context to streaming scenarios \cite{kim2018improved,yu2021dualmode}. These methods have applied KD both at the sequence level \cite{Pang18,panchapagesan2020efficient}, and the frame-level \cite{Chebotar16,mosner2019KD}. The early works on sequence level KD \cite{kim2016sequencelevel,kim2018improved} used a two-step procedure. However, a recently proposed method by Panchapagesan \etal \cite{panchapagesan2020efficient} allows for single-step co-distillation in RNNT models. Yu \etal~\cite{yu2021dualmode} used this loss function for training encoder modules capable of working in both streaming and full-context speech recognition scenarios. Wu \etal \cite{wu2021dynamic} applied the sequence level KD to train models of different sparsity levels. Their method results in unstructured sparsity that needs specialized implementation to fully exploit the computational benefit of sparsity. Our method varies the number of layers in the encoders to be able to reuse existing implementations for easy deployment. The methods \cite{Chebotar16,mosner2019KD} which perform frame-level KD, are usually for acoustic encoders used in a hybrid system. These methods are also a two-step procedure that requires a trained teacher model. Our proposed method is a single-step procedure with frame-level co-distillation and uses a shared predictor and joiner modules for implicitly sharing the sequence level knowledge between the acoustic encoders.

Co-distillation based collaborative learning methods \cite{lan2018knowledge,song2018collaborative,Guo_2020_CVPR} have been used for image classification. Lan \etal \cite{lan2018knowledge} proposed the idea of using a multi-branch network with shared low-level layers and training the branches using co-distillation. We apply a similar concept and combine it with the auxiliary task idea of Liu \etal \cite{liu2020auxtasks} for frame-level co-distillation.

Configurable neural networks are a class of models that are trained once and deployed in different configurations. There are two categories of such methods. In the first category~\cite{yu2018slimmable, yu2019UniversalSlimmable}, an appropriate network setting is used for inference based on a predetermined computational budget. In the second category, the methods train networks to dynamically adjust their resources during inference. The second category can be further divided based on whether the methods allow an external agent to control the resources during inference. Some methods~\cite{veit2018convolutional,wang2018skipnet,wu2021dynamic} can adjust the forward pass pathway in the network to reduce the computation based on the input. Another set of methods called \textit{anytime inference} \cite{huang2018multiscale,ruiz2020anytime} allow an external agent to stop the computation at any point and get the best possible prediction. Our method falls into the first category, where an appropriate acoustic encoder is selected based on a specific device's constraint.

\section{Collaborative Training of Encoders}

\subsection{Low Latency Emformer Transducer}
We focus on building models for low latency streaming on-device speech recognition using the Emformer \cite{shi2020emformer}  transducer. Emformer is an efficient extension of the Augmented Memory Transformer (AM-TRF) \cite{Wu2020}. They both perform streaming Transformer~\cite{vaswani_attention} based speech recognition by splitting an utterance into multiple segments, and decoding a segment along with the context of surrounding segments. 
The model size and the computation cost of an Emformer encoder are determined by the input dimension, size of the feed-forward network, and the number of layers. We demonstrate a training methodology for a group of Emformer encoders of different depths.

\begin{figure}[t]
    \centering
    \includegraphics[height=4in]{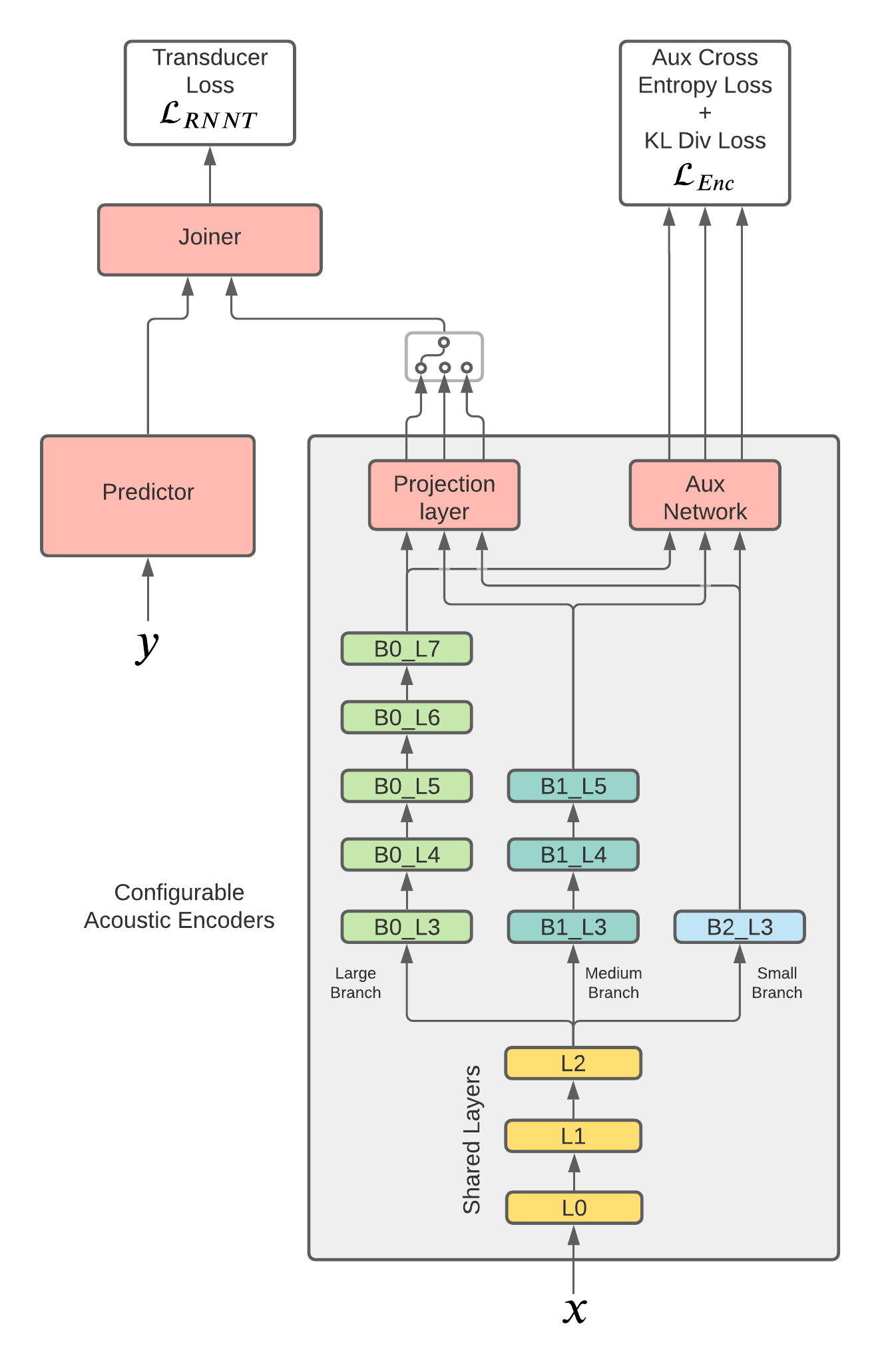}
    \caption{Block diagram of the training setup for the collaborative acoustic encoders. The nodes in yellow are the shared layers between the branches. $B_0$, $B_1$ and $B_2$ are three different encoders with different sizes. }
    \label{fig:collabTS}
    \vspace{-6pt}
\end{figure}

\subsection{Collaborative Training}
As shown in Figure \ref{fig:collabTS}, we design a group of Emformer encoders with a shared set of layers $F$ that then splits into multiple branches $\{B_0,B_1,\dots,B_{n-1}\}$ depending on the desired number of encoders based on the size. The number of layers in the shared network is denoted by $L_{F}$, and the number of layers in a branch $B_i$ is denoted by $L_{B_i}$. For deployment and decoding, we extract a copy of the shared layers and only one branch of the encoder based on the specific device requirement. Hence, the effective number of layers during the decoding is $L_{F}+L_{B_i}$ for a model created from branch $B_i$.

Figure \ref{fig:collabTS} shows that the predictor and the joiner network are shared with all the encoder branches. Given the input acoustic feature sequence $\bm{x}=\{x_1,\dots,x_T\}$ with sequence length $T$ and target token sequence $\bm{y}=\{y_1,\dots,y_U\}$ with length $U$ where $y_u \in\mathcal{Y}$, we can get the embedding representation $\bm{h}^x_i$ for each encoder $B_i$ and $\bm{h}^y$ for the predictor as follows:
\begin{align}
&\bm{h}^x_i = q(f_{i}(\bm{x})) \\
&\bm{h}^y = g(\{\varnothing,\bm{y}\}).
\end{align} where $\varnothing$ is a blank token. $f_i$ and $g$ are the projections from the acoustic encoder $B_i$ and the predictor, respectively. $q$ is the shared projection layer for each acoustic encoder. By combining each encoder's hidden representation with the predictor's output, the joiner generates the logits for each encoder $B_i$.
\begin{align}
&\bm{l}_i = z(\bm{h}^x_i, \bm{h}^y). \label{encoder}
\end{align}
where each element $l_{(t,u),i}$ in $\bm{l}_i$ is the logit given the acoustic sequence $\{x_1,...,x_{t}\}$ and label sequence $\{\varnothing, y_{1},...,y_u\}$. We can get the transducer loss~\cite{graves2012sequence} for each acoustic encoder $B_i$ by passing the logits through a softmax function and applying the forward and backward algorithm. We use the sum of all the encoders' transducer loss as the final transducer loss for the sequence transducer model with multiple acoustic encoders.
\begin{equation}
    \mathcal{L}_{Trans} = - \sum_{i=0}^{n-1} \log P_i(\bm{y}|\bm{x})
\end{equation}
The term $P_i(\bm{y}|\bm{x})$ is the summation of all the alignment probabilities $P_i(\bm{a}|\bm{x})$ where $\bm{a}$ is an alignment with elements $a_t \in \{\mathcal{Y} \cup \varnothing\}$. The removal of blank tokens from $\bm{a}$ gives the target sequence $\bm{y}$. In the work of Liu~ \etal\cite{liu2020auxtasks}, the outputs from the intermediate layers of an acoustic encoder are also connected to the transducer loss through a joiner. However, its purpose is to stabilize the deep encoder training where the gradients are passed back to the encoder without updating the weights of the joiner using the intermediate transducer losses. In contrast to their work, we update the joiner weights using the transducer losses obtained from the embeddings of all the encoder branches.

The proposed collaborative training method also includes a shared auxiliary network that projects the last layer outputs from each of the branches to logits for predicting the frame-level chenone \cite{le2019senones} targets. The auxiliary network consists of a hidden layer with RELU activation followed by an output layer. The same auxiliary network ($\text{MLP}_{\text{aux}}$) is used for all the branches. The output from the auxiliary network is passed through a softmax operator that produces a probability distribution $P_i(\bm{s}|\bm{x})$ over the chenone targets given the embedding representation from an encoder branch $B_i$. It is defined as
\begin{align}
P_i(\bm{s}|\bm{x})= \mathrm{softmax}(\text{MLP}_{\text{aux}}(f_{i}(\bm{x}))
\end{align}
where $\bm{s}$ is the frame level sequence of chenone targets.

We apply the cross-entropy loss for chenone prediction from each branch and also a KL divergence loss between pairs of branches. Since the number of pairs can blow up combinatorially, we identify the largest encoder branch and create pairs with each of the remaining branches. The largest branch plays a role as a teacher in knowledge distillation. The loss for the outputs from the encoder branches can be written as
\begin{align}
    & \mathcal{L}_{CE} = - \sum_{i=0}^{n-1} \log P_i(\bm{s}|\bm{x}) \label{eqn:ce}  \\ 
    & \mathcal{L}_{KL} = - \sum_{i\ne K} P_K(\bm{s}|\bm{x}) \log \frac{P_i(\bm{s}|\bm{x})}{P_K(\bm{s}|\bm{x})} \label{eqn:kl} \\
    & \mathcal{L}_{Enc} =  \mathcal{L}_{CE} + \mathcal{L}_{KL}
    \label{eqn:L_enc}
\end{align}
where $K$ is the index for branch $B_K$ with the largest size. $\mathcal{L}_{CE}$ in Eqn~\ref{eqn:ce} is the cross entropy loss with the groundtruth and $\mathcal{L}_{KL}$ in Eqn~\ref{eqn:kl} is the KL divergence loss with the soft targets from the largest branch. 
The entire model with all the branches is trained jointly in an end-to-end manner and the overall loss function is given by
\begin{equation}
    \mathcal{L} = \sum_{\mathcal{D}} \left( \mathcal{L}_{Trans} + \lambda \mathcal{L}_{Enc}\right)
    \label{eqn:total_loss}
\end{equation}
where $\lambda$ is a hyperparameter to control for the weight given to the encoder loss and $\mathcal{D}$ represents the training dataset.

\section{Experiments}
\subsection{Datasets and Setup}
We perform experiments using the LibriSpeech corpus \cite{librispeech}.
The LibriSpeech corpus contains around 960 hours of training data and around 5 hrs each of ``dev-clean", ``dev-other", ``test-clean", ``test-other" datasets. 

The input acoustic features are 80-dimensional log Mel filter bank energies. The width of an audio frame is 25ms with a stride of 10ms. We augment the data using speed perturbation \cite{ko2015audio} and SpecAugment \cite{Park2019} without the time warping modifier. The 80 dimensional features for each audio frame are projected to a 128 dimensional vector and the frames are concatenated at a stride of 4 to form a 512 dimensional vector. The input dimension of each Emformer layer is 512, the number of heads in the self-attention layer is 8 and the size of the feed-forward network is 2048. A final projection layer outputs a 1024 dimensional vector to pass to the joiner. We use a dropout of 0.1 in the self-attention and the feed-forward layers. Since we perform experiments for low latency conditions, the center chunk size and look-ahead context size in Emformer are set to 160ms and 40ms, respectively. The algorithmic latency~\cite{shi2020emformer} of the acoustic encoders is 120 ms.

The predictor is a three layer LSTM with an input and hidden dimension of 512 and the output is projected to 1024 dimensions before passing it to the joiner. The joiner adds the acoustic and label embeddings and projects it to the target space of word pieces obtained using SentencePiece \cite{sentencepiece}. For experiments which use the auxiliary task, the frame level chenone \cite{le2019senones} targets are generated by forced alignment using a hybrid HMM-GMM system. This hybrid model was bootstrapped using the standard Kaldi \cite{Kaldi} LibriSpeech recipe.

We use the Alignment Restricted RNNT \cite{ar-rnnt} loss for training the models. All the models are trained for 120 epochs using a learning rate of 1e-3 with the ADAM optimizer. The learning rate is increased linearly from 2e-7 to 1e-3 over 10K warmup updates for all the experiments. We use the last checkpoint for performing decoding.

We experiment with acoustic encoders of different number of layers. The total number of parameters including the predictor and the joiner can be found in Table \ref{tab:layers_params}.
\begin{table}[t]
\caption{Number of parameters in the Speech Recognition model based on the number of layers in the acoustic encoder.}
\centering
\begin{tabular}{|l|c|c|c|c|c|}
\hline
\textbf{Num Layers} & 20 & 18 & 14 & 10 & 7 \\ \hline
\textbf{Num Params (M)} & 76.7 & 70.4 & 57.8 & 45.2 & 35.7 \\ \hline
\end{tabular}
\label{tab:layers_params}
\vspace{-10pt}
\end{table}

\subsection{Results}

We compare the performance of the collaboratively trained models with the models trained separately. We use a weight of 0.6 for the auxiliary cross-entropy loss when training the models individually based on the observation from the work~\cite{liu2020auxtasks}. When training the models collaboratively, we use a weight of 0.1 for the auxiliary cross-entropy loss and the distillation loss. Table \ref{tab:lambda_results} shows the word error rate results based on collaborative training method using 3 different encoders with the different weight values ($\lambda$ in Eqn. \ref{eqn:total_loss}) used during the collaborative training. We observe no single value that performs consistently well across all the model sizes and the dataset partitions. A value of 0.05 gives better results on the test-other partition, and a value of 0.2 gives better results on the test-clean partition for the 20 and 14 layers models. Finally, we use the value of 0.1 since it performs better on average across the model sizes.

\begin{table}[h]
\centering
\caption{Word error rate (WER) results from using different weight value for the encoder loss during collaborative training. The two numbers in each cell are the WER results on test-clean and test-other datasets, respectively.}
\label{tab:lambda_results}
\begin{tabular}{c|c|r|r|}
\cline{2-4}
\multicolumn{1}{l|}{} & \multicolumn{3}{c|}{\textbf{Num Layers}} \\ \hline
\multicolumn{1}{|c|}{\textbf{\begin{tabular}[c]{@{}c@{}}$\bm{\lambda}$\end{tabular}}} & \textbf{20} & \multicolumn{1}{c|}{\textbf{14}} & \multicolumn{1}{c|}{\textbf{7}} \\ \hline
\multicolumn{1}{|c|}{0.05} & \multicolumn{1}{r|}{3.47, \textbf{9.14}} & 3.78, \textbf{9.67} & \textbf{4.84}, 12.80 \\ \hline
\multicolumn{1}{|c|}{0.1} & \multicolumn{1}{r|}{3.50, 9.23} & 3.57, 9.80 & 4.88, \textbf{12.21} \\ \hline
\multicolumn{1}{|c|}{0.2} & \textbf{3.38}, 9.19 & \multicolumn{1}{l|}{\textbf{3.56}, 9.74} & \multicolumn{1}{l|}{4.90, 13.41} \\ \hline
\multicolumn{1}{|c|}{0.3} & \multicolumn{1}{r|}{3.43, 9.21} & 3.60, 9.68 & 5.00, 13.18 \\ \hline
\end{tabular}
\end{table}

\begin{table*}[t]
\centering
\caption{Word error rate (WER) results from models trained with and without the different factors of the collaborative learning method. We also include results for collaboratively trained models with different number of branches. The two numbers in each cell are the WER results on test-clean and test-other datasets respectively. Bold values indicate the best performing model in a group when compared to the baselines. All the encoders have the same algorithmic latency of 120ms with 160ms chunk size and 40ms look-ahead context size.}
\label{tab:main_results}
\resizebox{\textwidth}{!}{%
\begin{tabular}{cccc|r|r|r|r|r|}
\cline{5-9}
\multicolumn{1}{l}{} & \multicolumn{1}{l}{} & \multicolumn{1}{l}{} & \multicolumn{1}{l|}{} & \multicolumn{5}{c|}{\textbf{Num Layers}} \\ \hline
\multicolumn{1}{|c|}{\textbf{\textbf{\begin{tabular}[c]{@{}c@{}}Shared Predictor \\ and Joiner\end{tabular}}}} & \multicolumn{1}{c|}{\textbf{\textbf{\begin{tabular}[c]{@{}c@{}}Shared Encoder \\ Layers\end{tabular}}}} & \multicolumn{1}{c|}{\textbf{\begin{tabular}[c]{@{}c@{}}Aux task with \\ CE loss\end{tabular}}} & \textbf{\begin{tabular}[c]{@{}c@{}}Aux task with \\ CE+KLDiv Loss\end{tabular}} & \multicolumn{1}{c|}{\textbf{20}} & \multicolumn{1}{c|}{\textbf{18}} & \multicolumn{1}{c|}{\textbf{14}} & \multicolumn{1}{c|}{\textbf{10}} & \multicolumn{1}{c|}{\textbf{7}} \\ \hline
\multicolumn{1}{|c|}{-} & \multicolumn{1}{c|}{-} & \multicolumn{1}{c|}{-} & - & 3.76, 9.86 & 3.83,  9.82 & 3.87, 10.35 & 4.29, \textbf{11.31} & 4.65, 12.50 \\ \hline
\multicolumn{1}{|c|}{-} & \multicolumn{1}{c|}{-} & \multicolumn{1}{c|}{y} & - & 3.58, 10.48 & 3.80, 10.91 & 3.94, 11.29 & 4.19, 11.76 & 4.75, 12.81 \\ \hline
\multicolumn{4}{|l|}{} & \multicolumn{5}{c|}{\textbf{4 branches}} \\ \hline
\multicolumn{1}{|c|}{y} & \multicolumn{1}{c|}{-} & \multicolumn{1}{c|}{-} & - & 4.18, 10.46 &  4.35, 10.55 & 4.52, 11.06 & 4.97, 12.01 & \multicolumn{1}{c|}{-} \\ \hline
\multicolumn{1}{|c|}{y} & \multicolumn{1}{c|}{-} & \multicolumn{1}{c|}{y} & - & 3.90, 9.61 & 3.97, 9.92 & 4.25, 10.95 & 4.53, 11.68 & \multicolumn{1}{c|}{-} \\ \hline
\multicolumn{1}{|c|}{y} & \multicolumn{1}{c|}{-} & \multicolumn{1}{c|}{y} & \multicolumn{1}{c|}{y} & 3.83, 9.72 & 3.87, 9.69 & 4.10, 10.40 & 4.53, 11.33 & \multicolumn{1}{c|}{-} \\ \hline
\multicolumn{1}{|c|}{y} & \multicolumn{1}{c|}{y} & \multicolumn{1}{c|}{-} & - & 3.69, 9.95 & 3.83, 10.16 & 3.96, 10.34 & 4.42, 11.85 & \multicolumn{1}{c|}{-} \\ \hline
\multicolumn{1}{|c|}{y} & \multicolumn{1}{c|}{y} & \multicolumn{1}{c|}{y} & - & 3.52, 9.39 & 3.47, \textbf{9.49} & \textbf{3.70}, 10.24 & \textbf{4.02}, 11.53 & \multicolumn{1}{c|}{-} \\ \hline
\multicolumn{1}{|c|}{y} & \multicolumn{1}{c|}{y} & \multicolumn{1}{c|}{y} & y & \textbf{3.45, 9.34} & \textbf{3.39},  9.57 & 3.72, \textbf{10.10} & 4.17, 11.57 & \multicolumn{1}{c|}{-} \\ \hline
\multicolumn{4}{|c|}{} & \multicolumn{5}{c|}{\textbf{3 branches}} \\ \hline
\multicolumn{1}{|c|}{y} & \multicolumn{1}{c|}{-} & \multicolumn{1}{c|}{y} & y & \textbf{3.38,  8.87} & \multicolumn{1}{c|}{-} & \textbf{3.57,  9.77} & \multicolumn{1}{c|}{-} & \textbf{4.38, 11.84} \\ \hline
\multicolumn{1}{|c|}{y} & \multicolumn{1}{c|}{y} & \multicolumn{1}{c|}{y} & y & 3.50,  9.23 & \multicolumn{1}{c|}{-} & 3.57,  9.80 & \multicolumn{1}{c|}{-} & 4.88, 12.21 \\ \hline
\multicolumn{4}{|c|}{} & \multicolumn{5}{c|}{\textbf{2 branches}} \\ \hline
\multicolumn{1}{|c|}{y} & \multicolumn{1}{c|}{y} & \multicolumn{1}{c|}{y} & y & 3.49,  9.36 & \multicolumn{1}{c|}{-} & 3.72,  9.95 & \multicolumn{1}{c|}{-} & \multicolumn{1}{c|}{-} \\ \hline

\end{tabular}%
}
\vspace{-6pt}
\end{table*}

The baseline models in our experiments are the ones trained separately. The first row of Table~\ref{tab:main_results} shows the word error rate (WER) results from the baseline model trained using the transducer loss without  any auxiliary task. The second row of Table~\ref{tab:main_results} shows the WER from different models trained using transducer loss together with the frame-level cross-entropy loss based on the chenone alignment. 
We observe that the performance is mixed when using the auxiliary task. This could indicate that the original weight for the auxiliary loss proposed in~\cite{liu2020auxtasks} might have been sub-optimal and requires additional tuning.

There are four critical factors in the proposed collaborative training method: shared predictor and joiner, shared low-level encoder layers, an auxiliary task with cross-entropy (CE) loss, and the auxiliary task with CE and KL divergence loss. We perform an ablation study of these factors with a model of four different encoder branches. We also experiment with a different number of branches ranging from two to four. When using the setting with shared encoder layers, the first six layers are common between the different encoder branches. We show the results grouped by the number of branches in Table \ref{tab:main_results} for the experiments done using collaborative training. 

The collaborative training results with four branches show that the performance incrementally improves as we enable the different factors. The WER of the model that uses all the four factors of our proposed method improves by 3-11\% and 2.5-5.5\% on the test-clean and test-other partitions, respectively. The last two rows in the group using four branches differ in whether they use the KL divergence loss for the auxiliary task or not. The results indicate that the performance obtained from these two settings is very close to each other, but they are still significantly better than their baseline counterparts. 

One of the interesting observations is that the performance of the largest branch sees an improvement along with that of the student branches. This can be attributed to the shared modules which are improved due to the training on the inputs from the different branches. An outlier result is that the ten layers model's performance did not show an improvement on the test-other partition. 

Similar to the trend with the four encoder branches, the model with two encoder branches trained collaboratively by enabling all the factors also achieves significantly better results than the baseline models. The last row in Table \ref{tab:main_results} shows that the 20 layers branch gets a relative WER reduction over 7\% on test-clean and 5\% on test other, compared with a baseline model of the same size. The 14 layers branch gets a similar WER as the 20 layers baseline model with a 25\% reduction in model size.

We observed that when training a collaborative set of three branches, the seven layers model results were worse than the baseline. Hence, we experimented with a setting that did not include the shared low-level encoder layers. Interestingly, this setting provided better performance than the other settings used for the three branches. The results show that without using layer sharing in encoders, the 20 layers model trained with collaborative learning achieves the best  WER among all the cases, with over 10\% improvement on the test-clean and test-other partitions. However, the training time as measured using the train wall time from the fairseq~\cite{ott2019fairseq} logs is 28\% slower when there are no shared layers between the different encoders.

\section{Conclusions}
We have presented a collaborative learning method to train a sequence transducer model with multiple acoustic encoders for low latency on-device ASR. 
Our method can train models of different sizes at once and allows picking one specific acoustic encoder for deployment that meets the computational constraint of a device. 
The proposed collaborative learning method takes advantage of shared predictor and joiner modules along with shared low level encoder layers to improve the performance and reduce the training time.
A context-dependent graphemic state prediction task was used to provide the transducer model with forced alignment information that acted as a bridge for transferring knowledge among different acoustic encoders. 
The experiments on the LibriSpeech dataset based on a low latency constraint showed that the collaborative learning method can improve each acoustic encoder compared with training them separately. 
The large encoder got over 10\% relative WER reduction on both test-clean and test-other evaluation datasets. 
The collaboratively trained smaller encoders also got 3-11\% relative WER reduction on both evaluation sets when compared with the smaller baseline models.

\section{Acknowledgements}
\vspace{-3pt}
We thank Yuan (June) Shangguan and Jay Mahadeokar for their help in setting up the experiments and discussions regarding the datasets and methods in this paper.

\bibliographystyle{IEEEtran}
\bibliography{references}

% Generated by IEEEtran.bst, version: 1.13 (2008/09/30)
\begin{thebibliography}{10}
\providecommand{\url}[1]{#1}
\csname url@samestyle\endcsname
\providecommand{\newblock}{\relax}
\providecommand{\bibinfo}[2]{#2}
\providecommand{\BIBentrySTDinterwordspacing}{\spaceskip=0pt\relax}
\providecommand{\BIBentryALTinterwordstretchfactor}{4}
\providecommand{\BIBentryALTinterwordspacing}{\spaceskip=\fontdimen2\font plus
\BIBentryALTinterwordstretchfactor\fontdimen3\font minus
  \fontdimen4\font\relax}
\providecommand{\BIBforeignlanguage}[2]{{%
\expandafter\ifx\csname l@#1\endcsname\relax
\typeout{** WARNING: IEEEtran.bst: No hyphenation pattern has been}%
\typeout{** loaded for the language `#1'. Using the pattern for}%
\typeout{** the default language instead.}%
\else
\language=\csname l@#1\endcsname
\fi
#2}}
\providecommand{\BIBdecl}{\relax}
\BIBdecl

\bibitem{Wu2020}
C.~Wu, Y.~Wang, Y.~Shi, C.-F. Yeh, and F.~Zhang, ``{Streaming Transformer-Based
  Acoustic Models Using Self-Attention with Augmented Memory},'' in
  \emph{INTERSPEECH}, 2020.

\bibitem{shi2020emformer}
Y.~Shi, Y.~Wang, C.~Wu, C.-F. Yeh, J.~Chan, F.~Zhang, D.~Le, and M.~Seltzer,
  ``Emformer: Efficient memory transformer based acoustic model for low latency
  streaming speech recognition,'' in \emph{ICASSP}, 2021.

\bibitem{gulati2020conformer}
A.~Gulati, J.~Qin, C.-C. Chiu, N.~Parmar, Y.~Zhang, J.~Yu, W.~Han, S.~Wang,
  Z.~Zhang, Y.~Wu, and R.~Pang, ``Conformer: Convolution-augmented transformer
  for speech recognition,'' in \emph{INTERSPEECH}, 2020.

\bibitem{Hinton_KD}
G.~Hinton, O.~Vinyals, and J.~Dean, ``Distilling the knowledge in a neural
  network,'' in \emph{NeurIPS Deep Learning and Representation Learning
  Workshop}, 2015.

\bibitem{zhang2018deep}
Y.~Zhang, T.~Xiang, T.~M. Hospedales, and H.~Lu, ``Deep mutual learning,'' in
  \emph{CVPR}, 2018.

\bibitem{anil18codistillation}
R.~Anil, G.~Pereyra, A.~Passos, R.~Ormandi, G.~E. Dahl, and G.~E. Hinton,
  ``Large scale distributed neural network training through online
  distillation,'' in \emph{ICLR}, 2018.

\bibitem{lan2018knowledge}
X.~Lan, X.~Zhu, and S.~Gong, ``Knowledge distillation by on-the-fly native
  ensemble,'' in \emph{NeurIPS}, 2018.

\bibitem{graves2012sequence}
A.~Graves, ``Sequence transduction with recurrent neural networks,'' in
  \emph{Representation Learning Workshop, ICML}, 2012.

\bibitem{rao2017asru}
K.~Rao, H.~Sak, and R.~Prabhavalkar, ``Exploring architectures, data and units
  for streaming end-to-end speech recognition with rnn-transducer,'' in
  \emph{ASRU}, 2017.

\bibitem{he2018streaming}
Y.~He, T.~N. Sainath, and et~al., ``Streaming end-to-end speech recognition for
  mobile devices,'' in \emph{ICASSP}, 2019.

\bibitem{zhang2020transformer}
Q.~Zhang, H.~Lu, H.~Sak, A.~Tripathi, E.~McDermott, S.~Koo, and S.~Kumar,
  ``Transformer transducer: A streamable speech recognition model with
  transformer encoders and rnn-t loss,'' in \emph{ICASSP}, 2020.

\bibitem{vaswani_attention}
A.~Vaswani, N.~Shazeer, N.~Parmar, J.~Uszkoreit, L.~Jones, A.~N. Gomez, L.~u.
  Kaiser, and I.~Polosukhin, ``Attention is all you need,'' in \emph{NeurIPS},
  2017.

\bibitem{le2019senones}
D.~Le, X.~Zhang, W.~Zheng, C.~Fügen, G.~Zweig, and M.~L. Seltzer, ``From
  senones to chenones: Tied context-dependent graphemes for hybrid speech
  recognition,'' in \emph{ASRU}, 2019.

\bibitem{liu2020auxtasks}
C.~Liu, F.~Zhang, D.~Le, S.~Kim, Y.~Saraf, and G.~Zweig, ``Improving rnn
  transducer based asr with auxiliary tasks,'' in \emph{SLT}, 2020.

\bibitem{librispeech}
V.~{Panayotov}, G.~{Chen}, D.~{Povey}, and S.~{Khudanpur}, ``Librispeech: An
  asr corpus based on public domain audio books,'' in \emph{ICASSP}, 2015.

\bibitem{Chebotar16}
Y.~Chebotar and A.~Waters, ``Distilling knowledge from ensembles of neural
  networks for speech recognition,'' in \emph{INTERSPEECH}, 2016.

\bibitem{Pang18}
R.~Pang, T.~Sainath, R.~Prabhavalkar, S.~Gupta, Y.~Wu, S.~Zhang, and C.-C.
  Chiu, ``Compression of end-to-end models,'' in \emph{INTERSPEECH}, 2018.

\bibitem{panchapagesan2020efficient}
S.~Panchapagesan, D.~S. Park, C.-C. Chiu, Y.~Shangguan, Q.~Liang, and
  A.~Gruenstein, ``Efficient knowledge distillation for rnn-transducer
  models,'' in \emph{ICASSP}, 2021.

\bibitem{li2017KD}
J.~Li, L.~M. Seltzer, X.~Wang, R.~Zhao, and Y.~Gong, ``Large-scale domain
  adaptation via teacher-student learning,'' in \emph{INTERSPEECH}, 2017.

\bibitem{Asami17}
T.~{Asami}, R.~{Masumura}, Y.~{Yamaguchi}, H.~{Masataki}, and Y.~{Aono},
  ``Domain adaptation of dnn acoustic models using knowledge distillation,'' in
  \emph{ICASSP}, 2017.

\bibitem{Manohar18}
V.~{Manohar}, P.~{Ghahremani}, D.~{Povey}, and S.~{Khudanpur}, ``A
  teacher-student learning approach for unsupervised domain adaptation of
  sequence-trained asr models,'' in \emph{SLT}, 2018.

\bibitem{Meng19}
Z.~Meng, J.~Li, Y.~Gaur, and Y.~Gong, ``Domain adaptation via teacher-student
  learning for end-to-end speech recognition,'' in \emph{ASRU}, 2019.

\bibitem{mosner2019KD}
L.~Mošner, M.~Wu, A.~Raju, S.~H.~K. Parthasarathi, K.~Kumatani, S.~Sundaram,
  R.~Maas, and B.~Hoffmeister, ``Improving noise robustness of automatic speech
  recognition via parallel data and teacher-student learning,'' in
  \emph{ICASSP}, 2019.

\bibitem{kim2018improved}
S.~Kim, M.~L. Seltzer, J.~Li, and R.~Zhao, ``Improved training for online
  end-to-end speech recognition systems,'' in \emph{INTERSPEECH}, 2018.

\bibitem{yu2021dualmode}
J.~Yu, W.~Han, A.~Gulati, C.-C. Chiu, B.~Li, T.~N. Sainath, Y.~Wu, and R.~Pang,
  ``Dual-mode asr: Unify and improve streaming asr with full-context
  modeling,'' in \emph{ICLR}, 2021.

\bibitem{kim2016sequencelevel}
Y.~Kim and A.~M. Rush, ``Sequence-level knowledge distillation,'' in
  \emph{EMNLP}, 2016.

\bibitem{wu2021dynamic}
Z.~Wu, D.~Zhao, Q.~Liang, J.~Yu, A.~Gulati, and R.~Pang, ``Dynamic sparsity
  neural networks for automatic speech recognition,'' in \emph{ICASSP}, 2021.

\bibitem{song2018collaborative}
G.~Song and W.~Chai, ``Collaborative learning for deep neural networks,'' in
  \emph{NeurIPS}, 2018.

\bibitem{Guo_2020_CVPR}
Q.~Guo, X.~Wang, Y.~Wu, Z.~Yu, D.~Liang, X.~Hu, and P.~Luo, ``Online knowledge
  distillation via collaborative learning,'' in \emph{CVPR}, 2020.

\bibitem{yu2018slimmable}
J.~Yu, L.~Yang, N.~Xu, J.~Yang, and T.~Huang, ``Slimmable neural networks,'' in
  \emph{ICLR}, 2018.

\bibitem{yu2019UniversalSlimmable}
J.~Yu and T.~Huang, ``Universally slimmable networks and improved training
  techniques,'' in \emph{ICCV}, 2019.

\bibitem{veit2018convolutional}
A.~Veit and S.~Belongie, ``Convolutional networks with adaptive inference
  graphs,'' in \emph{ECCV}, 2018.

\bibitem{wang2018skipnet}
X.~Wang, F.~Yu, Z.-Y. Dou, T.~Darrell, and J.~E. Gonzalez, ``Skipnet: Learning
  dynamic routing in convolutional networks,'' in \emph{ECCV}, 2018.

\bibitem{huang2018multiscale}
G.~Huang, D.~Chen, T.~Li, F.~Wu, L.~van~der Maaten, and K.~Q. Weinberger,
  ``Multi-scale dense networks for resource efficient image classification,''
  in \emph{ICLR}, 2018.

\bibitem{ruiz2020anytime}
A.~Ruiz and J.~Verbeek, ``Anytime inference with distilled hierarchical neural
  ensembles,'' in \emph{AAAI}, 2021.

\bibitem{ko2015audio}
T.~Ko, V.~Peddinti, D.~Povey, and S.~Khudanpur, ``Audio augmentation for speech
  recognition,'' in \emph{INTERSPEECH}, 2015.

\bibitem{Park2019}
D.~S. Park, W.~Chan, Y.~Zhang, C.-C. Chiu, B.~Zoph, E.~D. Cubuk, and Q.~V. Le,
  ``{SpecAugment: A Simple Data Augmentation Method for Automatic Speech
  Recognition},'' in \emph{INTERSPEECH}, 2019.

\bibitem{sentencepiece}
T.~Kudo and J.~Richardson, ``{S}entence{P}iece: A simple and language
  independent subword tokenizer and detokenizer for neural text processing,''
  in \emph{EMNLP: System Demonstrations}, 2018.

\bibitem{Kaldi}
D.~Povey, A.~Ghoshal, G.~Boulianne, N.~Goel, M.~Hannemann, Y.~Qian, P.~Schwarz,
  and G.~Stemmer, ``The kaldi speech recognition toolkit,'' in \emph{ASRU},
  2011.

\bibitem{ar-rnnt}
J.~Mahadeokar, Y.~Shangguan, D.~Le, G.~Keren, H.~Su, T.~Le, C.-F. Yeh,
  C.~Fuegen, and M.~L. Seltzer, ``Alignment restricted streaming recurrent
  neural network transducer,'' in \emph{SLT}, 2021.

\bibitem{ott2019fairseq}
M.~Ott, S.~Edunov, A.~Baevski, A.~Fan, S.~Gross, N.~Ng, D.~Grangier, and
  M.~Auli, ``fairseq: A fast, extensible toolkit for sequence modeling,'' in
  \emph{NAACL-HLT: Demonstrations}, 2019.

\end{thebibliography}

\end{document}